\documentclass{article}

\usepackage{microtype}
\usepackage{graphicx}
\usepackage{subfigure}
\usepackage{booktabs} 
\usepackage{placeins}
\usepackage{diagbox}
\usepackage{multirow}
\usepackage{anyfontsize}
\usepackage{comment}
\usepackage{breqn}
\usepackage{textcomp}
\usepackage{makecell}

\usepackage{hyperref}

\usepackage[accepted]{icml2024}

\usepackage{amsmath}
\usepackage{amssymb}
\usepackage{mathtools}
\usepackage{amsthm}

\usepackage[capitalize,noabbrev]{cleveref}

\theoremstyle{plain}

\theoremstyle{definition}

\theoremstyle{remark}

\usepackage[textsize=tiny]{todonotes}

\icmltitlerunning{FlowDA: Flow Matching for Weather-scale Data Assimilation}

\usepackage{times}
\usepackage{bm}

\makeatletter
\newcommand{\HEADER}[1]{\ALC@it\underline{\textsc{#1}}\begin{ALC@g}}
\newcommand{\ENDHEADER}{\end{ALC@g}}
\makeatother

\renewcommand{\algorithmicrequire}{\textbf{Input:}}
\renewcommand{\algorithmicensure}{\textbf{Output:}}

\renewcommand{\headrulewidth}{0pt}

\usepackage{natbib}
\usepackage{hyperref}
\def\equationautorefname~#1\null{(#1)\null}
\def \d {\mathrm{d}}
\def \lp {\left(}
\def \rp {\right)}

\def \bu {\mathbf{u}}
\def \bx {\mathbf{x}}
\def \bX {\mathbf{X}}
\def \by {\mathbf{y}}
\def \bz {\mathbf{z}}

\def \bpsi {\bm{\psi}}

\def \bX {\mathbf{X}}

\def \bxFM {\bz}

\def \bxa {\bx^{\mathrm{a}}} 
\def \bxb {\bx^{\mathrm{b}}} 
\def \bxg {\bx^{\mathrm{g}}} 
\def \bxo {\bx^{\mathrm{o}}} %
\def \bxobs {\bxo} 
 
\def \veltrue {\bu_{\tau}}

\def \density {\rho^{\mathrm{o}}}                 
\def \densityvector {{\bm{\rho}}^{\mathrm{o}}_t}     

\def \bbR {\mathbb{R}} 
\def \bbE {\mathbb{E}} 
\def \bbRobs{\mathbb{R}^{M \times V}}               
\def \bbRstate {\mathbb{R}^{H \times W \times V}}   

\def \Unif {\mathrm{Unif}}

\def \mcalH {\mathcal{H}}

\begin{document}

\twocolumn[
\icmltitle{FlowDA: Accurate, Low-Latency Weather Data Assimilation via Flow Matching
}

\begin{icmlauthorlist}
\icmlauthor{Ran Cheng}{nus}
\icmlauthor{Lailai Zhu}{nus}
\end{icmlauthorlist}

\icmlaffiliation{nus}{Department of Mechanical Engineering, National University of Singapore}

\icmlcorrespondingauthor{Lailai Zhu}{lailai\_zhu@nus.edu.sg}

\icmlkeywords{Data Assimilation, Denoising Diffusion Model}

\vskip 0.3in
]

\printAffiliationsAndNotice{}

\begin{abstract}
Data assimilation (DA) is a fundamental component of modern weather prediction, yet it remains a major computational bottleneck in machine learning (ML)-based forecasting pipelines due to reliance on traditional variational methods. Recent generative ML-based DA methods offer a promising alternative but typically require many sampling steps and suffer from error accumulation under long-horizon auto-regressive rollouts with cycling assimilation.
We propose FlowDA, a low-latency weather-scale generative DA framework based on flow matching. FlowDA conditions on observations through a SetConv-based embedding and fine-tunes the Aurora foundation model to deliver accurate, efficient, and robust analyses. 
Experiments across observation rates decreasing from $3.9\%$ to $0.1\%$ demonstrate superior performance of FlowDA over strong baselines with similar tunable-parameter size. FlowDA further shows robustness to observational noise and stable performance in long-horizon auto-regressive cycling DA. Overall, FlowDA points to an efficient and scalable direction for data-driven DA.

\end{abstract}

\section{Introduction}
Weather forecasts and climate projections are critical for safeguarding lives, supporting decision-making, and improving resilience to extremes~\citep{ecmwf2020european}. 
For such tasks among other Earth-system modeling efforts, data assimilation plays a foundational role, supplying initial conditions for numerical weather prediction (NWP)~\cite{bauer2015quiet,documentation2020part} 
and generating observation-constrained reanalyses, e.g.,  the fifth-generation reanalysis (ERA5)~\citep{hersbach2020era5} from the European Centre for Medium-Range Weather Forecasts (ECMWF).

Conceptually, DA updates a numerical model forecast (background) towards an optimal estimate (analysis) by incorporating real-world observations from sources such as satellite sensors, in-situ stations, and radar, etc~\cite{Lei2025Overview1,Kalnay2002Overview2,Zou2025Overview3}. Mathematically, operational DA is commonly formulated in a Bayesian framework and implemented via variational methods or ensemble Kalman filters~\citep{bannister2017VarEnKFReview}. The former is typified by three-dimensional variational assimilation (3D-Var)~\citep{parrish19923DVar} and its more widely used extension, four-dimensional variational assimilation (4D-Var)~\citep{courtier19944dvar}.

Despite their theoretical rigor, these conventional DA approaches require solving high-dimensional optimization problems repeatedly, making them computationally intensive and thus a persistent bottleneck in operations. For example, the overall DA within ECMWF's operations constitutes approximately $40\%$ of their computational budget. 
The reminder is largely allocated for NWP---currently being increasingly eclipsed by the proliferation of machine-learning-based weather prediction (MLWP)~\cite{bi2023panguweathe,chen2023fengwu,chen2023fuxi,lam2023graphcast,price2023gencast,Kochkov2024NeuralGCM,alet2025FGN,bonev2025fourcastnet3}
and AI Earth-system foundation models~\cite{nguyen2023climax,schmude2024prithvi,bodnar2025aurora}. 
These AI-driven frameworks deliver comparable or superior forecast skill relative to NWP at an orders-of-magnitude lower inference cost. This dramatic acceleration of the prediction phase further exposes the relative inefficiency of traditional DA, motivating drastically faster machine-learning-powered DA (MLDA) solutions. 
Beyond efficiency, AIDA enables generating initial conditions for MLWP directly from raw observations~\citep{gupta2026healda}. Despite their growing performance advantages over NWPs, MLWP models still rely on initial conditions provided by ERA5 or other analysis products derived from traditional NWP-DA workflows~\citep{dunstan2025fastnet}.

Early MLDA studies use ML to accelerate or enhance specific modules within a DA workflow, while retaining the core DA architecture. 
Recent work has pursued end-to-end forecasting
pipelines that alternate MLWP and MLDA in an iterative cycling procedure conditioned on raw observations~\cite{xiao2024vaevar, vandal2024earthnet}. Besides, \citet{fan2025physicallyconsistent} proposes latent DA (LDA)---a Bayesian DA in a latent space learned from multivariate global atmospheric data via an autoencoder. Their research demonstrates that latent space assimilation improves analysis quality and forecast skill over traditional model-space DA in both idealized and real-world settings. Furthermore, HealDA~\citep{gupta2026healda} serves as a standalone raw-observation-to-state DA module independent of background fields, enabling the initialization of general MLWP models without fine-tuning. Another major thread in MLDA focuses on developing generative DA~\cite{rozet2023scoreDA, andrychowicz2023metnet3, qu2024deepgenerativeda, chen2024fnp}---exclusively harnessing diffusion models~\cite{ho2020DDPM, karras2022EDM}---to infer analysis fields conditioned on observational data. Representative examples include DiffDA~\cite{Huang2024DiffDA}, Appa~\cite{andry2025appa}, and LO-SDA~\cite{sun2025losda}. Despite their success, diffusion-based generative models can be computationally expensive at inference due to their multi-step sampling procedure, creating an efficiency bottleneck for generative MLDA. This motivates us to explore Flow Matching (FM)~\cite{lipman2023FM, esser2024rectified, le2023voicebox,yim2023fast}---a modern variant of diffusion model---as a route to faster generative DA. 
This choice is grounded that FM can match diffusion-level performance using substantially fewer sampling steps.

\begin{figure*}[h!]
    \centering
    \includegraphics[width=0.9\linewidth]{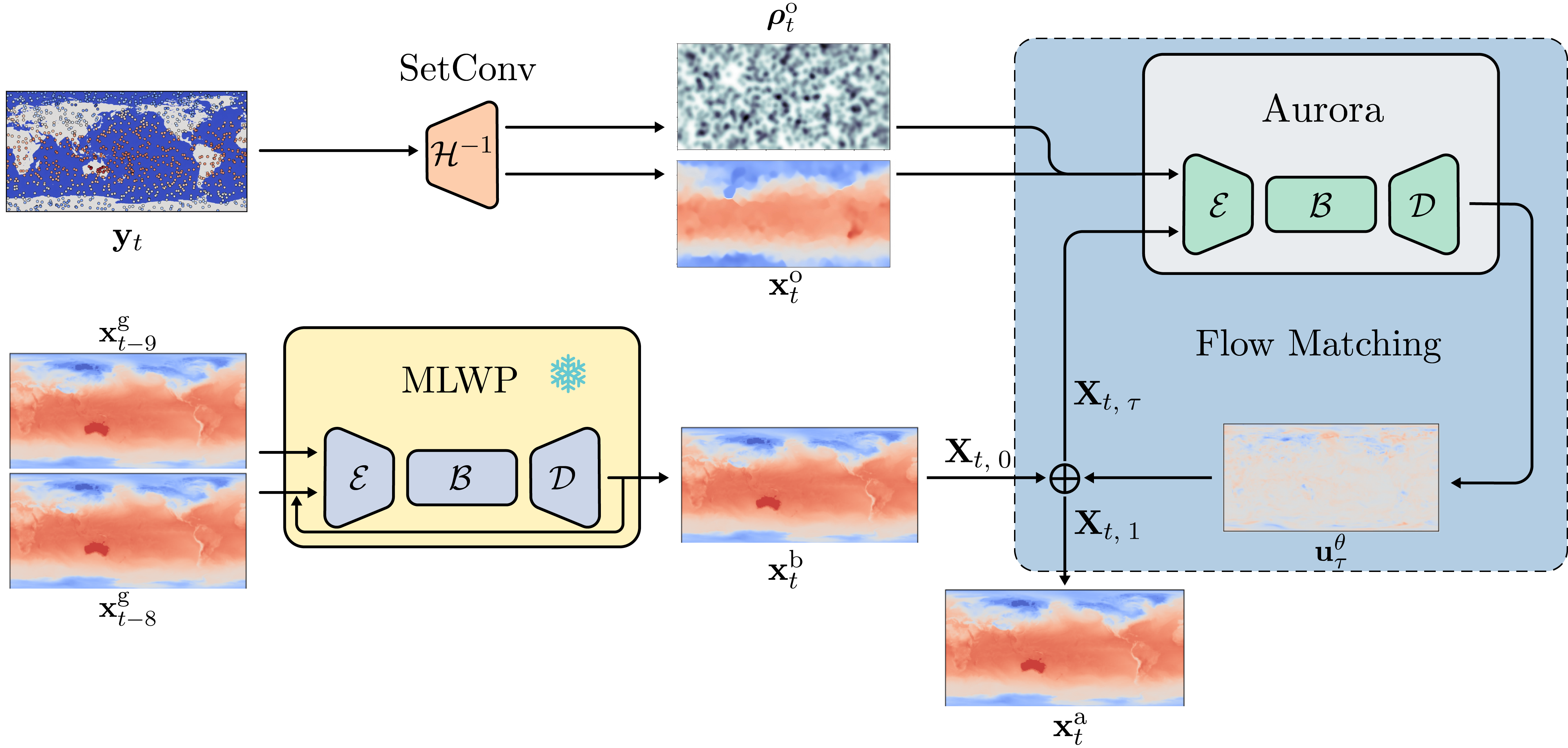}
    \caption{
    Workflow of FlowDA inference for single-step DA initialized from a 48-hour forecast. FlowDA first applies a SetConv layer whose weights depend on the local observation density $\alpha_m$ and relative distance, serving as an analogue of inverse observation operator $\mathcal{H}^{-1}$. A fine-tuned Aurora model then takes the flow state ${\mathbf{X}}_{t,\tau}$ conditioned on ${\mathbf{x}}^{\mathrm{o}}_t$ and ${\bm{\rho}}^{\mathrm{o}}t$, and estimates a velocity field ${\mathbf{u}}^{\theta}{\tau}$ that corrects the background ${\mathbf{x}}^{\mathrm{b}}_t$ into the analysis ${\mathbf{x}}^{\mathrm{a}}_t$. The analysis is produced through iterative forward Euler integration.
    }
    \label{fig:diagram_model}
\end{figure*}

In this work, we present FlowDA, an FM-based generative MLDA framework for accurate and efficient weather-scale DA. FlowDA is scaled to the global atmospheric system at $0.25^{\circ}$ resolution and 6-hour intervals, covering  4 surface fields and 5 atmospheric variables across 13 pressure levels. By integrating the Earth-system foundation model Aurora~\citep{bodnar2025aurora} and fine-tuning it for DA, FlowDA leverages Aurora’s pre-trained atmospheric dynamics and its learned physical structure.

FlowDA effectively corrects background states under severely limited observational coverage and exhibits strong robustness to observational noise. 
On multiple DA benchmarks, FlowDA surpasses existing MLDA baselines, including the purely data-driven DiffDA~\citep{Huang2024DiffDA} and the ML-augmented variational approach VAE-Var~\citep{xiao2024fengwu4dvar}---the state-of-the-art (SOTA) model for long-horizon cycling DA. FlowDA improves both analysis accuracy and inference efficiency (when comparable) while maintaining a similar or smaller trainable-parameter budget.

\section{Methods}
\subsection{Problem Statement}\label{sec:prob_formulation}
We cast data assimilation as a Bayesian inference problem. At discrete time $t$, we estimate the atmospheric state $\bxa_t$ (the analysis) from a prior gridded forecast $\bxb_t \in \bbR^{H \times W \times V}$ (the background) and $M$ sparse observations $\by_t \in \bbR^{M \times V }$. Here, $H \times W \times V $ denotes a stacked multivariate atmospheric state comprising surface variables and pressure-level fields across multiple vertical levels, discretized on an $H \times W$ latitude-longitude grid for each of $V$ variables.
Specifically, we compute the maximum-a-posteriori estimate $\bxa_t$ by expanding the probability density function $p$ as   
\begin{align}
    \bxa_t
    & = \arg\max_{\bx_t} \, p(\bx_t \mid \bxb_t,\;\by_t),
\end{align}
where we have assumed the independence between the background errors and observation errors as commonly done~\citep{bouttier1999data}.

\subsection{Model Overview}

We propose FlowDA, a 
flow-matching  
framework for global weather-scale data assimilation. 
FlowDA delivers low-latency inference and lower analysis error than prevailing baselines, with a comparable or smaller number of trainable parameters. 
FlowDA comprises principal components: 1) a SetConv-based learned observation embedding and 2) a fine-tuned AI foundation model, Aurora. 
The former maps an observation field $\by_t$ to a gridded state estimate $\bxobs_t \in \bbR^{H \times W \times V}$ and an associated measurement-density field $\densityvector$; the latter conditions on this estimate to transform the background $\bxb_t$ into the analysis $\bxa_t$.
This formulation decouples observation optimality from assimilation dynamics, enabling modular exploitation of observational information without compromising the physical continuity of the background.

\subsection{Flow Matching Framework}
Flow matching (FM)~\citep{lipman2023FM} is a generative modeling framework that requires no additional noise scheduling and enables fast sampling, 
making it widely applied in generative tasks such as images, audio, and proteins~\cite{esser2024rectified,le2023voicebox, yim2023fast}. Recently, FM has been adopted for weather forecast~\citep{couairon2024archesweather,he2025flowcast}, DA for high-dimensional dynamical systems tested on model configurations~\citep{transue2025flow}.
We will briefly introduce the working principle of FM below.

Let $\bxFM \in \bbR^{d}$ denote a state variable. FM defines a probability path $p_{\tau}$ indexed by a pseudo-time $\tau \in [0,1]$, which evolves from a source distribution $p_{0}$ to a target distribution $p_{1}$. Here, $p_{\tau}$ denotes a probability density on $\bbR^{d}$ such that $\bxFM_{\tau} \sim p_{\tau}$. FM introduces a time-dependent \textit{flow} function $\bpsi_{\tau}$ such that $\bpsi_{\tau} \lp \bxFM_0 \rp = \bxFM_{\tau}$. This flow is governed by the ordinary differential equation (ODE)
\begin{align}\label{eq:continuity}
    \frac{\d \bpsi_{\tau} \lp \bxFM_0 \rp }{\d {\tau}} = \veltrue \lp\bm{\psi}_{\tau} \lp \bxFM_0 \rp\rp,
\end{align}
where $\veltrue$ represents the marginal velocity  quantifying the rate of change of the state $\bxFM$ with respect to the generative time $\tau$. 
Typically, FM parameterizes the velocity field with a neural network $\veltrue^{\theta}$ and learns it from samples by minimizing the conditional FM  loss:
\begin{align}
\mathcal{L}_{\rm CFM}\lp\theta \rp = \bbE_{\tau  \sim \Unif \lp 0, 1 \rp , \bxFM_{\tau}, \bxFM_1 } || \veltrue^{\theta}\lp \bxFM_{\tau} \rp - \veltrue \lp \bxFM_{\tau} | \bxFM_{1} \rp  ||^2.
\end{align}
Under a straight-line probability  path from $\bxFM_{0}\sim p_{0}$ to $\bxFM_{1}\sim p_{1}$, the conditional velocity $\veltrue \lp \bxFM_{\tau} | \bxFM_{1} \rp$ simplifies to $\bxFM_{1}-\bxFM_{0}$, yielding
\begin{align}\label{eq:fm_loss}
\mathcal{L}_{\rm CFM}\lp\theta \rp = \bbE_{\tau  \sim  \Unif \lp 0, 1 \rp , \bxFM_{0}, \bxFM_1 } || \veltrue^{\theta}\lp \bxFM_{\tau} \rp - \lp \bxFM_1 - \bxFM_0 \rp  ||^2.
\end{align}
At inference, FM draws $\bxFM_{0}\sim p_{0}$ and integrates Eq.~\autoref{eq:continuity} over $\tau\in[0,1]$ using the learned field $\veltrue^{\theta}$ to obtain a terminal sample $\bxFM_{1}\sim p_{1}$, which can be computed with standard ODE solvers such as forward Euler.

\subsection{FlowDA: Flow Matching for Weather-scale Data Assimilation}\label{sec:cond_pred}
FlowDA achieves weather-scale DA via conditional FM by learning a linear probability path from the background distribution  to the analysis distribution. 
Following FM, we denote the background state at discrete time $t$ as $\bxb_t = \bX_{t,0}$ when $\tau=0$, the analysis state as $\bxg_t = \bX_{t,1}$ when $\tau=1$, and the intermediate flow state at generative time $\tau$ as $\bX_{t,\tau} \sim p_{\tau}$.

Because the analysis field is conditioned on the observation field $\by_t$, the conditional FM loss for FlowDA is
\begin{dmath}\label{eq:fm_da_loss}
\mathcal{L}_{\rm CFM}\lp\theta \rp = \bbE_{\tau  \sim  \Unif \lp 0, 1 \rp, t, \bX_{t, 0}, \bX_{t, 1} } || \bu^{\theta}_{t,\tau} \left[ \bX_{t, \tau}, \mcalH^{-1}\lp \by_t \rp \right] - \lp \bX_{t,1} - \bX_{t,0} \rp  ||^2,
\end{dmath}
where $t$ is uniformly drawn from the training time window. Here,  $\mcalH^{-1}\lp \by_t \rp$ resembling an inverse observation operation, maps discrete measurements to a continuous gridded field, as detailed in \autoref{sec:obs}.

In the inference, we use 
a forward Euler integrator with 
a generative time step of $\delta \tau = 1/32$
to solve Eq.~\autoref{eq:continuity}:
\begin{align}
\bX_{t,\tau+\delta \tau} = \bX_{t,\tau} + \delta \tau \,\bu^{\theta}_{t,\tau} 
\end{align}
Compared with diffusion models that typically require hundreds of denoising steps, FlowDA enables substantially faster inference (see \autoref{para:single_DA}).

\subsection{Conditioning for Sparse Observations}\label{sec:obs}

Consistent with classical DA, FlowDA introduces an observational field as a conditioning signal that serves as an anchor. Typically, this field $\by_t \in \bbRobs$ consists of unstructured, sparse, and irregularly spaced observations, where the total number of observations $M$ and their sampling locations vary over time $t$. Integrating such an asynchronous, irregular field into a background state with fixed dimensionality is a fundamental challenge in DA.

To address this challenge, the prior AIDA framework DiffDA employs inpainting, while FNP adopts a neural-process approach, where the observation-state transformation is coupled to assimilation.
In contrast, FlowDA incorporates a module analogous to the inverse of 3D-Var observation operator $\mathcal{H}$, thereby decoupling the two processes. 

To ease the demonstration of FlowDA, we use synthetic observations. Specifically, the  observation field $\by_t \in \bbR^{M \times V}$ contains $V$ meteorological variables sampled at $M$ randomly selected locations from a gridded ground-truth dataset $\bxg_t \in \bbR^{H \times W \times V}$ at time $t$, the ERA5 reanalysis chosen here (see \autoref{sec:traindata} for details). Note that, though the spatial resolution ($H\times W$) of our forecast $\bxb_t$ matches that of ERA5 in this study, such a match is not a requirement of FlowDA.

Despite the structured synthetic observations used here, FlowDA is designed to assimilate irregularly spaced real measurements. To map an irregular observation field $\by_t$ into the model state space, we employ a $\text{SetConv}$ layer~\citep{Gordon2020ConvCNP} as $\mcalH^{-1}$ to 
lift the observations to a gridded representation, $\bxo_t = \text{SetConv}(\by_t) \in \bbRstate$.
Notably,  SetConv as adopted in Aardvark~\citep{Allen2025Endtoend} is particularly suitable for DA because it effectively lifts irregularly located discrete measurements into a continuous, gridded representation.

In practice, $\text{SetConv}(\by_t)$ constructs $\bxo_t$ on a per-grid-point basis. For the $n$-th grid point ($n \in [0, N-1]$ with $N = HW$), the lifted representation $\bxo_t|^{n}$ is computed as a normalized weighted sum over all $M$ observations:
\begin{equation}
\bxo_t|^{n}
= \frac{1}{\density_t|^{n}} \sum_{m=1}^{M} \phi_{mn} \, \by_t^{m},
\end{equation}
where $\phi_{mn}$ weights the influence of the $m$-th observation ($\by_t^m$) on the $n$-th grid. The normalization
$\density_t|^{n} = \sum_{m=1}^{M} \phi_{mn}$
represents the local observation density around this grid.

Let $\langle h^n, w^n \rangle$ denote the latitude and longitude of grid point $n$, and $\langle h^m, w^m \rangle$ those of observation $m$. We calculate the weight $\phi_{mn}$ via an anisotropic Multilayer Perceptron (MLP) kernel ~\citep{lowery2025mlpkernel},
\begin{equation}
\phi_{mn}
= \text{MLP}_h \left( h^n - h^m, \alpha_m \right)\cdot \, \text{MLP}_w \left( w^n - w^m, \alpha_m \right),
\end{equation}
where $\alpha_m$ denotes the local observation rate associated with the $m$-th observation (see~\autoref{sec:loc_obv}). Conditioning the kernel on $\alpha_m$ enables adaptive weighting tailored to irregular observation fields.

\begin{figure}[t!]
    \vskip -0.05in
    \centering
     \includegraphics[width=1\linewidth]{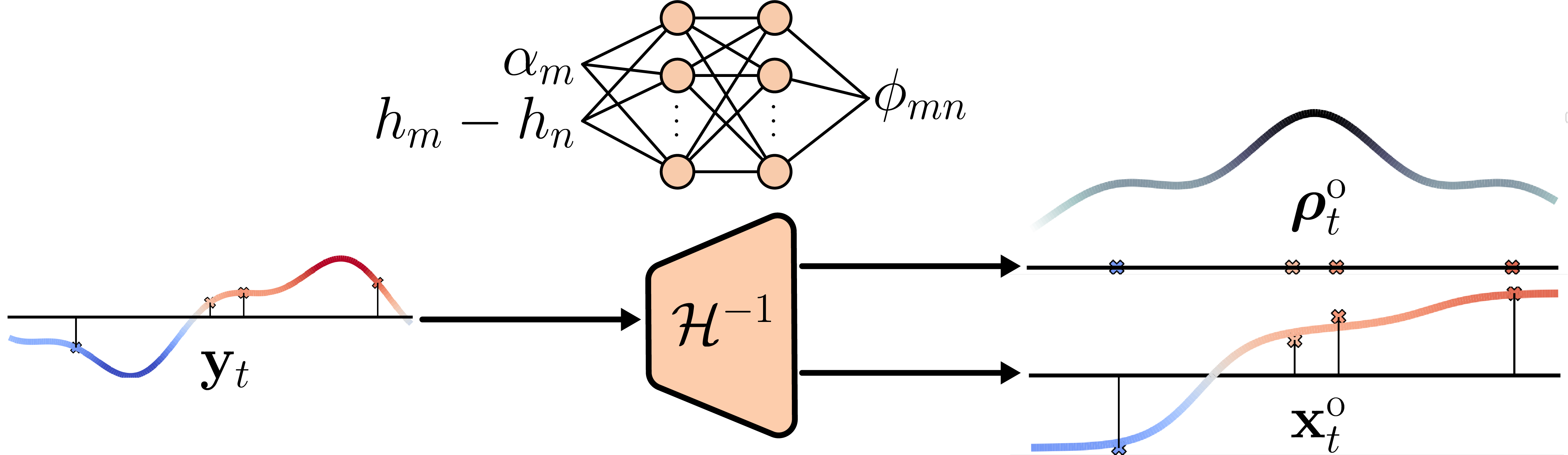}
    \caption{
    Workflow of the SetConv operator, which converts sparse observations $\by_t$ into a continuous model-space field $\bxobs_t$ and the corresponding observation density distribution $\densityvector$ via an MLP-based kernel.
    }
    \label{fig:setconv}
\end{figure}

In each flow-matching training step, gridded state estimate $\bxobs_t$, observation density distribution $\densityvector$ and the flow state $\bX_{t, \tau}$ jointly provided as inputs to learn a conditional velocity field. The resulting loss function is formulated as
\begin{equation}
    \mathcal{L}(\theta)
    = \mathbb{E}_{\tau, t} \lVert \bu^{\theta}_{t,\tau}(\bX_{t, \tau},\bxobs_t,\densityvector) - \veltrue(\bX_{t, \tau} \mid \bX_{t, 1}) \rVert^2
\end{equation}

\subsection{Adapting a Foundation Model for DA}
\paragraph{Aurora} 
Pretrained foundation models can be fine-tuned to solve downstream tasks with significantly lower training cost. Here, we achieve global weather-scale DA by fine-tuning a foundation model for the Earth system,  Aurora~\citep{bodnar2025aurora}.
Pretrained on large-scale geophysical datasets including ERA5~\citep{hersbach2020era5}, second-version Modern-Era Retrospective analysis for Research and Applications (MERRA-2)~\citep{gelaro2017MERRA2}, fourth-generation ECMWF Atmospheric Composition Reanalysis (EAC4)~\citep{inness2019EAC4}, and ECMWF High-Resolution Operational Forecast (HRES T0)~\citep{ecmwf_hres_2023},
Aurora employs a 3D Perceiver~\citep{jaegle2022perceiverio} encoder--decoder ($\mathcal{E}$--$\mathcal{D}$) with a 3D Swin Transformer~\citep{liu2021Swin} U-Net backbone ($\mathcal{B}$) to learn a general-purpose Earth-system representation with multiscale fidelity and temporal coherence. 
The base Aurora model and its lighter checkpoint, Aurora-Small, 
involve 1.3 B and 113 M parameters, respectively. 
Besides, we also use Aurora-Small-Forecast checkpoint as our MLWP model to obtain the background fields in this study.

\paragraph{Fine-tuning Aurora} \label{sec:fine-tuning}
As shown in \autoref{fig:diagram_model} and \autoref{eq:fm_da_loss}, we fine-tune Aurora to model the conditional marginal velocity $\bu^{\theta}_{t,\tau}$ from inputs $(\bX_{t,\tau}, \bxo_t, \density_t)$, trained by minimizing the conditional FM loss in Eq.~\eqref{eq:fm_da_loss}.
To achieve this goal, we employ
1) a parameter efficient Low-Rank Adaptation (LoRA) strategy and 2) full-parameter fine-tuning, yielding FlowDA-LoRA and FlowDA-Full, respectively. FlowDA-LoRA trains the encoder and decoder of the 1.3B-parameter base Aurora alongside additional rank-60 LoRA layers, involving 37M tunable parameters, while FlowDA-Full updates all 113M parameters of Aurora-Small.

\subsection{Inference}

Overall, the inference pipeline of FlowDA is detailed in Algorithm.\autoref{alg:inference}.
\begin{algorithm}
  \caption{Inference of FlowDA}
  \label{alg:inference}
  \begin{algorithmic}
  \REQUIRE background $\bxb_t$, sparse observations $\by_t$, flow step number $L$
  \ENSURE reanalysis field $\bxa_t$
  \STATE $\densityvector,\bxobs_t=\text{SetConv}(\by_t)$
  \STATE $\bX_{t, 0}=\bxb_t$
  \FOR{$\tau$ in $0,\;\frac{1}{L},\;\cdots,\;\frac{L-1}{L}$}
      \STATE $\bu^{\theta}_{t,\tau} = \text{Aurora}(\bX_{t, \tau},\,\bxobs_t,\,\densityvector)$
      \STATE $\delta\tau = \frac{1}{L}$ 
      \STATE $\bX_{t,\,\tau+\delta\tau} = \bX_{t, \tau} + \delta \tau\,\bu^{\theta}_{t,\tau}$
  \ENDFOR
  \STATE $\bxa_t=\bX_{t, 1}$
\end{algorithmic}
\end{algorithm}

\section{Experiments}

\subsection{Implementation}\label{sec:training}
\paragraph{Training dataset} \label {sec:traindata}

We use the ERA5 reanalysis dataset~\citep{hersbach2020era5} at $0.25 ^{\circ}$ spatial resolution over 1979--2015 
as the source of both ground truth $\bxg$ and synthetic observations $\by$ (see \autoref{sec:obs}). 
Our state comprises four surface variables, 2 m temperature (T2m), 10 m zonal (U10) and meridional (V10) winds, and mean sea level pressure (MSLP), alongside five upper-air variables (temperature, geopotential, zonal and meridional winds, and specific humidity) defined on 13 pressure levels. In total, this yields $V=69$ variables for each atmospheric column.

\paragraph{Training protocols}

In this study, training for FlowDA-LoRA denotes LoRA-based fine-tuning of the 1.3B-parameter base Aurora model, while that for FlowDA-Full means 
full-parameter fine-tuning of the 113M-parameter Aurora-Small variant (see \autoref{sec:fine-tuning}). 
Under both paradigms, We adopt a two-stage fine-tuning protocol.
In Stage I, we fine-tune Aurora for single-step 48-hour DA (one analysis update at lead time 48 h). 
Stage II warm-starts from the Stage-I checkpoint and continues fine-tuning for auto-regressive cycling DA, supporting multi-step analysis--forecast trajectories.

In Stage I for single-step DA, at each training iteration we sample a triplet $(\bxg_{t-9}, \bxg_{t-8}, \bxg_t)$ from the dataset,  representing ERA5 reanalysis states at three time points; here, each discrete time step equals 6 hours---a convention used throughout this work.
Meanwhile, the current-time groundtruth state 
$\bxg_t$ is randomly masked, retaining from 1,000 to 40,000 observations to construct the observation field $\by_t$. The corresponding global observation rates $\alpha$  are $\alpha \approx 0.1\%$ and $\alpha \approx 3.9\%$, respectively.
To construct the background at time $t$, we 
initialize Aurora-Small-Forecast with the two preceding states $(\bxg_{t-9}, \bxg_{t-8})$ 
and run an 8-step rollout (6 hours per step) to obtain $\bxb_t$. Although FlowDA can assimilate backgrounds produced by other MLWP or NWP systems, we adopt Aurora-Small-Forecast for MLWP throughout this study to preserve architectural consistency.

\begin{figure}[htp!]
    \centering
    \includegraphics[width=0.8\linewidth]{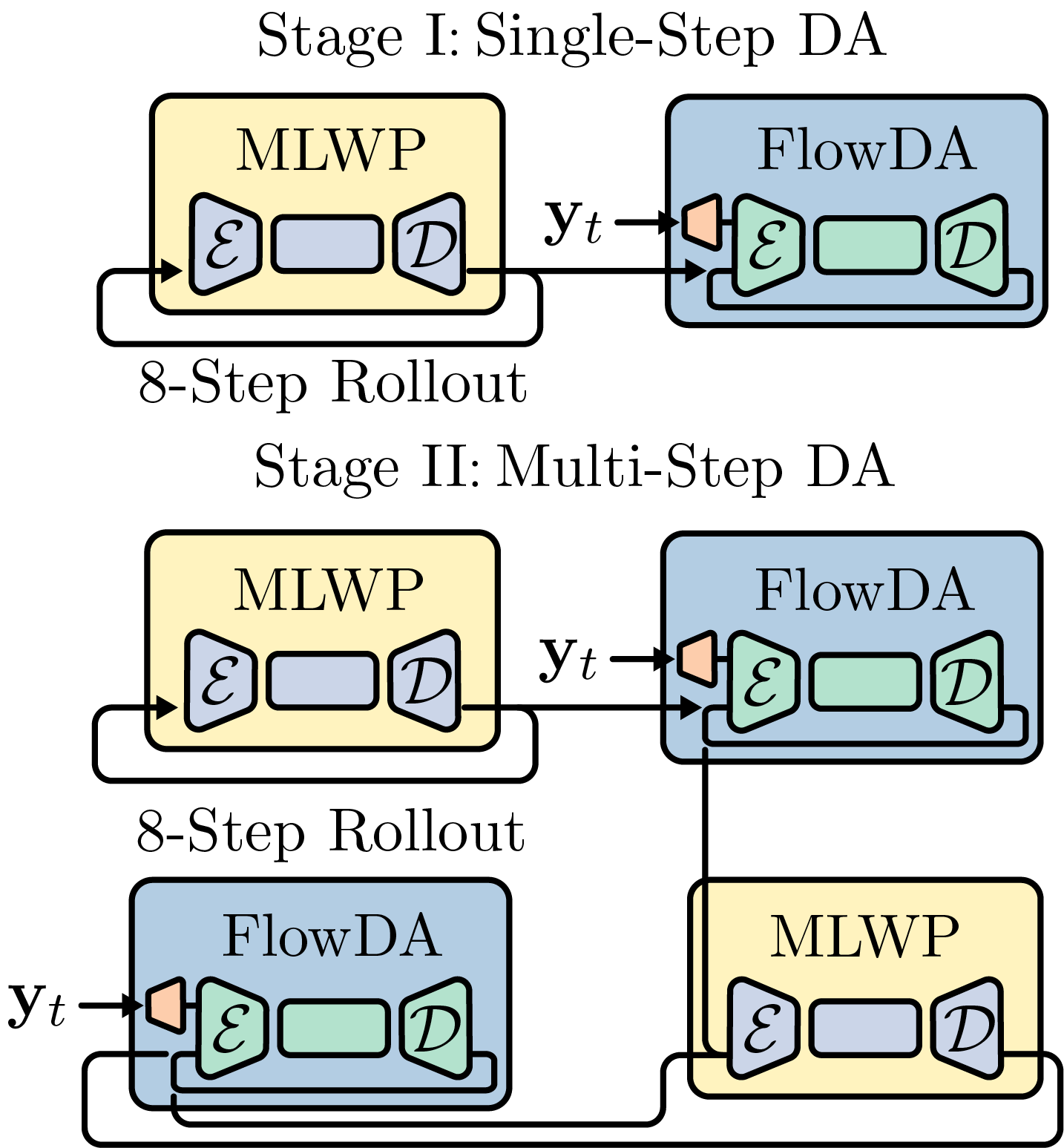}
    \caption{Stage I and II fine-tuning protocols.}
    \label{fig:Experiment_graph}
\end{figure}

In Stage II, we use multi-step fine-tuning to strengthen auto-regressive cycling DA. At each iteration, we sample a start time $t$ and a rollout length $T \sim \Unif(\left\{1,2,\ldots,8\right\})$, then alternate between (i) forecasting with Aurora-Small-Forecast to obtain the background and (ii) applying DA at every step (6-hour cadence) over the $T$-step trajectory.

Stage-I fine-tuning is performed for 12 epochs on 8 NVIDIA H200 GPUs with a per-GPU batch size of one. We use AdamW~\citep{loshchilov2018adamw} with a learning rate of $3\times 10^{-4}$ and enable activation checkpointing for memory efficiency. 
Under the same configuration, the resulting checkpoint is further fine-tuned for one additional epoch in Stage II to improve auto-regressive cycling DA.

\subsection{Experimental Settings and Results}
We test FlowDA on three canonical DA benchmarks: (1) single-step DA, (2) single-step DA with noisy observations, and (3) auto-regressive cycling DA. In all experiments, the background field is produced by Aurora-Small-Forecast from two consecutive ERA5 states. We assimilate ERA5-based synthetic observations with observation rates ranged in $0.1\% \lessapprox \alpha \lessapprox 3.9\%$. We evaluate FlowDA over the test period 01 January 2022 00 UTC through 16 January 2022 18 UTC, aiming for a consistent performance benchmark against DiffDA and VAE-Var.

We quantify the DA performance using the Root Mean Square Error (RMSE) of the analysis relative to the ground truth. We consider four surface variables (T2M, U10, V10, and MSLP) and five pressure-level variables: q100, u300, v300, z500, and t850. The raw magnitudes of these variables span a wide range, approximately $10^{-7}$ to $10^{5}$, underscoring the need to evaluate across heterogeneous scales~\cite{ashkboos2022ens10,rasp2024weatherbench2}.

\begin{table*}[thp!]
\centering
\caption{Single-step DA after a 48-hour forecast at time index $t$ (03 January 2022 06 UTC). 
Initiated from $t-8$, our forecast for FlowDA is obtained auto-regressively via Aurora-Small-Forecast.}
\label{tab:Single-step}
\vspace{0.254cm}
\fontsize{9pt}{9pt}\selectfont
\begin{tabular}{rlccccccccc}
\toprule
& & \multicolumn{4}{|c|}{Surface Variables} & \multicolumn{5}{c}{Pressure-level Variables} 
\\[-8pt]
\\ \cline{3-11} 
\\[-8pt]
& & \multicolumn{1}{|c}{T2M} & U10 & V10 & \multicolumn{1}{c|}{MSLP} & q100 & u300 & v300 & z500 & t850\\ 
&\textbf{Model} & \multicolumn{1}{|c}{(K)} & $(\mathrm{m}\,\mathrm{s}^{-1})$ & $(\mathrm{m}\,\mathrm{s}^{-1})$ & \multicolumn{1}{c|}{(hPa)} & ($\mathrm{g\,kg^{-1}}$) & ($\mathrm{m}\,\mathrm{s}^{-1}$) & ($\mathrm{m}\,\mathrm{s}^{-1}$) & ($\mathrm{m}^2\,\mathrm{s}^{-2}$) & (K) \\ 
\midrule
\\[-8pt]
\multicolumn{1}{l|}{\multirow{4}{*}{\rotatebox[origin=c]{90}{Forecast at}}} & \multicolumn{1}{l|}{$t-7$ $|$ FlowDA} & 0.64 & 0.60 & 0.60 & \multicolumn{1}{c|}{43.2} & 0.66e-7 & 1.38 & 1.45 & 32.5 & 0.52 \\[3pt]
\multicolumn{1}{l|}{}                  & \multicolumn{1}{l|}{$t-4$ $|$ FlowDA} & 0.96 & 1.00 & 0.97 & \multicolumn{1}{c|}{88.1} & 1.34e-7 & 2.32 & 2.31 & 61.8 & 0.79 \\[3pt]
\multicolumn{1}{l|}{}                  & \multicolumn{1}{l|}{$t$ (background) $|$ FlowDA} & 1.33 & 1.46 & 1.49 & \multicolumn{1}{c|}{150} & 1.89e-7 & 3.56 & 3.61 & 128 & 1.16 \\[3pt]
\multicolumn{1}{l|}{}                  & \multicolumn{1}{l|}{$t$ (background) $|$ DiffDA} & 1.27 & 1.26 & 1.28 & \multicolumn{1}{c|}{112} & 2.02e-7 & 2.88 & 2.98 & 93.2 & 1.00 \\
\\[-8pt]
\midrule
\\[-8pt]
\multicolumn{1}{l|}{\multirow{3}{*}{\rotatebox[origin=c]{90}{$\alpha$$\approx$3.9\%}}} & \multicolumn{1}{l|}{FlowDA-Full (113M)} & \textbf{0.54} & \textbf{0.50} & \textbf{0.50} & \multicolumn{1}{c|}{\textbf{25.2}} & \textbf{0.62e-07} & \textbf{0.91} & \textbf{0.91} & \textbf{14.6} & \textbf{0.41} \\[3pt]
\multicolumn{1}{c|}{}                  & \multicolumn{1}{l|}{FlowDA-LoRA (37M)} & 0.56 & 0.51 & 0.51 & \multicolumn{1}{c|}{26.6} & 0.67e-07 & 0.93 & 0.96 & 16.5 & 0.42 \\[3pt]
\multicolumn{1}{l|}{}                  & \multicolumn{1}{l|}{DiffDA (37 M)} & 0.58 & 0.66 & 0.65 & \multicolumn{1}{c|}{34.0} & 0.74e-07 & 1.02 & 1.07 & 19.5 & 0.43 \\
\\[-8pt]
\midrule
\\[-8pt]
\multicolumn{1}{l|}{\multirow{3}{*}{\rotatebox[origin=c]{90}{$\alpha\approx$1.0\%}}} & \multicolumn{1}{l|}{FlowDA-Full (113M)} & \textbf{0.71} & \textbf{0.72} & \textbf{0.71} & \multicolumn{1}{c|}{\textbf{34.5}} & \textbf{1.07e-7} & \textbf{1.47} & \textbf{1.41} & \textbf{19.4} & \textbf{0.58} \\[3pt]
\multicolumn{1}{l|}{}                  & \multicolumn{1}{l|}{FlowDA-LoRA (37M)} & 0.74 & 0.74 & 0.73 & \multicolumn{1}{c|}{37.6} & 1.14e-7 & 1.56 & 1.51 & 22.1 & 0.60 \\[3pt]
\multicolumn{1}{l|}{}                  & \multicolumn{1}{l|}{DiffDA (37 M)} & 0.93 & 1.06 & 1.05 & \multicolumn{1}{c|}{56.5} & 1.21e-7 & 2.00 & 1.97 & 37.2 & 0.70 \\
\\[-8pt]
\midrule
\\[-8pt]
\multicolumn{1}{l|}{\multirow{3}{*}{\rotatebox[origin=c]{90}{$\alpha\approx$0.4\%}}} & \multicolumn{1}{l|}{FlowDA-Full (113M)} & \textbf{0.84} & \textbf{0.90} & \textbf{0.89} & \multicolumn{1}{c|}{\textbf{44.2}} & \textbf{1.41e-7} & \textbf{1.96} & \textbf{1.93} & \textbf{27.0} & \textbf{0.70} \\[3pt]
\multicolumn{1}{l|}{}                  & \multicolumn{1}{l|}{FlowDA-LoRA (37M)} & 0.87 & 0.90 & 0.91 & \multicolumn{1}{c|}{47.6} & 1.44e-7 & 2.07 & 2.02 & 30.2 & 0.73 \\[3pt]
\multicolumn{1}{l|}{}                  & \multicolumn{1}{l|}{DiffDA (37 M)} & 1.21 & 1.35 & 1.30 & \multicolumn{1}{c|}{85.4} & 1.53e-7 & 2.77 & 2.86 & 65.9 & 0.92 \\
\\[-8pt]
\midrule
\\[-8pt]
\multicolumn{1}{l|}{\multirow{3}{*}{\rotatebox[origin=c]{90}{$\alpha\approx$0.1\%}}} & \multicolumn{1}{l|}{FlowDA-Full (113M)} & \textbf{1.02} & \textbf{1.17} & \textbf{1.17} & \multicolumn{1}{c|}{\textbf{68.1}} & 1.83e-7 & \textbf{2.76} & \textbf{2.71} & \textbf{51.8} & \textbf{0.87} \\[3pt]
\multicolumn{1}{l|}{}                  & \multicolumn{1}{l|}{FlowDA-LoRA (37M)} & 1.07 & 1.17 & 1.17 & \multicolumn{1}{c|}{74.4} & 1.83e-7 & 2.88 & 2.83 & 54.6  & 0.91 \\
\multicolumn{1}{l|}{}                  & \multicolumn{1}{l|}{DiffDA (37 M)} & 1.49 & 1.55 & 1.49 & \multicolumn{1}{c|}{111} & \textbf{\fbox{1.80e-7}} & 3.45 & 3.57 & 90.1 & 1.10 \\
\\[-8pt]
\bottomrule
\end{tabular}
\end{table*}

\paragraph{Single-Step DA}\phantomsection\label{para:single_DA}

At time index $t$ corresponding to 03 January 2022 06 UTC, we perform single-step DA at a 48-hour lead time, taking the 48-hour free-running forecast valid at $t$ as the background.

\autoref{tab:Single-step} summarizes the RMSEs of analyses produced by FlowDA-Full (113M), FlowDA-LoRA (37M), and DiffDA (37M) as the observation rate $\alpha$ decreases from $\approx 3.9\%$ to $\approx 0.1\%$. Here, 113M denotes the number of trainable parameters for FlowDA-Full, and the same convention applies to the other models.
For all models, the assimilation error drops monotonically with increasing $\alpha$.

For nearly all variables and observation rates, both FlowDA-Full and FlowDA-LoRA achieve lower analysis RMSE than DiffDA; the only exception is q100 at $\alpha\approx 0.1\%$, when the latter is slightly better. Besides, FlowDA-Full consistently outperforms FlowDA-LoRA, though with a marginal gap. 

Importantly, FlowDA yields robust assimilation: the analysis error 
remains consistently lower than the background error across all observation rates, including the most sparse setting $\alpha \approx 0.1\%$. In contrast, DiffDA fails to 
consistently improve upon the background when $\alpha \approx 0.1\%$.

While FlowDA-LoRA outperforms DiffDA in both accuracy and robustness, we emphasize three points. First, both share a comparable trainable-parameter budget (37M). Second, FlowDA-LoRA operates under a more challenging setting---starting from a larger background error (see the Forecast block of \autoref{tab:Single-step}). Third, FlowDA-LoRA offers markedly lower inference latency: it takes approximately 4 minutes, compared with about 15 minutes for DiffDA, both measured on a single NVIDIA A100 GPU. As a side note, FlowDA-Full is faster still, achieving roughly 2 minutes per inference despite involving 113M trainable parameters.

Moreover, at a high observation rate ($\alpha \approx 3.9\%$), FlowDA reduces error below that of the 6-hour ($t-7$) forecast across all variables, corresponding to an effective lead-time gain of 42 hours.

\begin{figure}[h!]
    \centering
    \includegraphics[width=1 \linewidth]{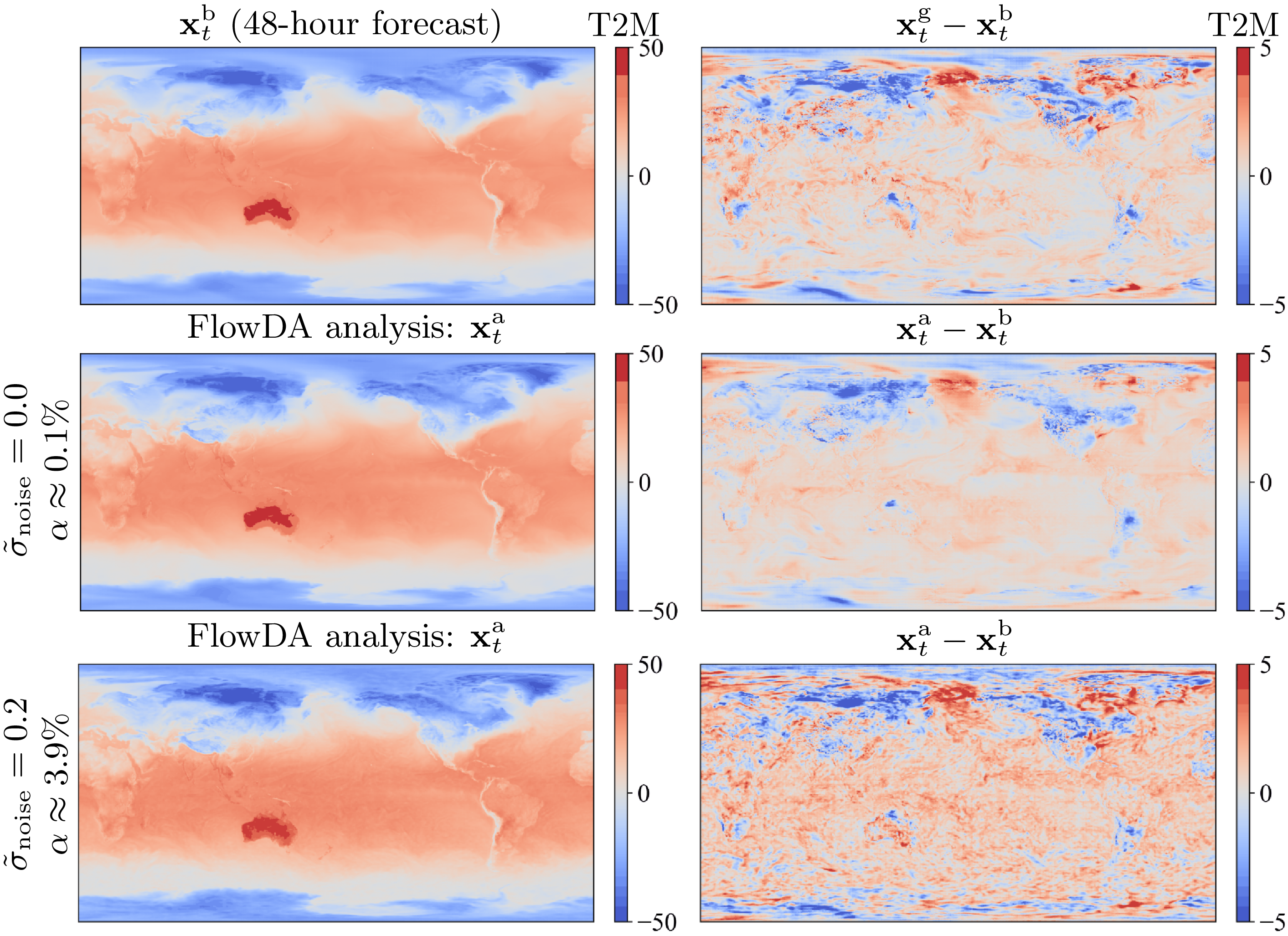}
    \caption{
Single-step DA with ERA5-based observations perturbed by additive Gaussian noise. Row 1: background field (left) from a 48-hour free-running forecast and the corresponding background error (right). Row 2: FlowDA analysis (left) and increment (right) for $\alpha \approx 0.1\%$ and $\tilde{\sigma}_{\text{noise}}=0.0$. Row 3: FlowDA analysis (left) and increment (right) for $\alpha \approx 3.9\%$ and $\tilde{\sigma}_{\text{noise}}=0.2$.
    }
    \label{fig:casestudy}
\end{figure}

\begin{table}[h!]
\centering
\caption{DA as in \autoref{tab:Single-step}, but with synthetic noisy observations---ERA5 reanalysis perturbed by additive zero-mean Gaussian noise.
}
\label{tab:obs_noisy}
\vspace{0.254cm}
\fontsize{9pt}{9pt}\selectfont
\begin{tabular}{rcccccc}
\toprule
& & \multicolumn{5}{|c}{Meteorological variables} \\ 
\\[-7pt]
\cline{3-7} 
\\[-5pt]
& & \multicolumn{1}{|c}{T2M} & U10 & q100 & z500 & t850 \\
& $\Tilde{\sigma}_{\text{noise}}$ & \multicolumn{1}{|c}{(K)} & ($\mathrm{m}\,\mathrm{s}^{-1}$) & ($\mathrm{g\,kg^{-1}}$) &  ($\mathrm{m}^{2}\,\mathrm{s}^{-2}$) & (K) \\
\midrule
\multicolumn{2}{c|}{Background}  & 1.33 & 1.26 & 1.89e-7 & 128 & 1.16 \\
\midrule
\parbox[t]{4mm}{\multirow{5}{*}{\rotatebox[origin=c]{90}{$\alpha$ $\approx$ 3.9\%}}}
\vline  & \multicolumn{1}{c|}{0.00}  & 0.54 & 0.49 & 0.61e-7 & 16.3 & 0.42 \\[3pt]
\vline  & \multicolumn{1}{c|}{0.05}  & 0.58 & 0.52 & 0.67e-7 & 18.5 & 0.44 \\[3pt]
\vline  & \multicolumn{1}{c|}{0.10}  & 0.70 & 0.58 & 0.79e-7 & 22.9 & 0.49 \\[3pt]
\vline  & \multicolumn{1}{c|}{0.20}  & 0.98 & 0.74 & 0.99e-7 & 35.5 & 0.60 \\[3pt]
\midrule
\parbox[t]{4mm}{\multirow{5}{*}{\rotatebox[origin=c]{90}{$\alpha$ $\approx$ 1.0\%}}}
\vline  & \multicolumn{1}{c|}{0.00}  & 0.71 & 0.72 & 1.07e-7 & 19.4 & 0.58 \\[3pt]
\vline  & \multicolumn{1}{c|}{0.05}  & 0.75 & 0.74 & 1.13e-7 & 22.9 & 0.60 \\[3pt]
\vline  & \multicolumn{1}{c|}{0.10}  & 0.85 & 0.81 & 1.21e-7 & 28.0 & 0.64 \\[3pt]
\vline  & \multicolumn{1}{c|}{0.20}  & 1.07 & 0.94 & 1.33e-7 & 39.3 & 0.73 \\[3pt]
\midrule
\parbox[t]{4mm}{\multirow{5}{*}{\rotatebox[origin=c]{90}{$\alpha$ $\approx$ 0.4\%}}}
\vline  & \multicolumn{1}{c|}{0.00}  & 0.84 & 0.90 & 1.41e-7 & 27.0 & 0.70 \\[3pt]
\vline  & \multicolumn{1}{c|}{0.05}  & 0.86 & 0.91 & 1.42e-7 & 29.4 & 0.72 \\[3pt]
\vline  & \multicolumn{1}{c|}{0.10}  & 0.92 & 0.96 & 1.47e-7 & 34.1 & 0.74 \\[3pt]
\vline  & \multicolumn{1}{c|}{0.20}  & 1.07 & 1.05 & 1.54e-7 & 44.1 & 0.81 \\[3pt]
\midrule
\parbox[t]{4mm}{\multirow{5}{*}{\rotatebox[origin=c]{90}{$\alpha$ $\approx$ 0.1\%}}}
\vline  & \multicolumn{1}{c|}{0.00}  & 1.02 & 1.17 & 1.83e-7 & 51.8 & 0.87 \\[3pt]
\vline  & \multicolumn{1}{c|}{0.05}  & 1.04 & 1.17 & 1.83e-7 & 53.5 & 0.87 \\[3pt]
\vline  & \multicolumn{1}{c|}{0.10}  & 1.07 & 1.18 & 1.83e-7 & 56.5 & 0.88 \\[3pt]
\vline  & \multicolumn{1}{c|}{0.20}  & 1.12 & 1.22 & 1.83e-7 & 65.7 & 0.92 \\[3pt]
\midrule
\multicolumn{2}{c|}{DiffDA} & \multirow{2}{*}{1.49} & \multirow{2}{*}{1.55} & \multirow{2}{*}{1.80e-7} & \multirow{2}{*}{90.1} & \multirow{2}{*}{1.10} \\
\multicolumn{2}{c|}{($\Tilde{\sigma}_{\text{noise}}=0$)} & & & & & \\

\bottomrule
\end{tabular}
\end{table}

\paragraph{Single-Step DA with Noisy Observations} 
\phantomsection\label{para:noisy_obs}
In practice, measurement data $\by_t$ are subject to observation error, in contrast to the ERA5-based synthetic observations used in  \autoref{para:single_DA}. To reflect this practical challenge, we introduce noise into the ERA5 observations. Concretely, we perturb each observed variable by adding zero-mean Gaussian noise, $\mathcal{N}(0,\sigma)$ with $\sigma=\tilde{\sigma}_{\text{noise}}\,\sigma_{\text{Aurora}}$. Here, $\sigma_{\text{Aurora}}$ is the standard deviation of that variable in the Aurora models; $\tilde{\sigma}_{\text{noise}}$ is a dimensionless parameter that sets the relative noise level for all perturbed variables, which ranges in $\tilde{\sigma}_{\text{noise}}\in[0, 0.2]$ in this work.
As a concrete example, the standard deviation
for T2M in Aurora is $\sigma_{\text{Aurora}} \approx 21.22,\mathrm{K}$. Setting $\tilde{\sigma}_{\text{noise}}=0.05$ then corresponds to additive noise with standard deviation $\sigma \approx 1.06$ K, comparable in magnitude to the assimilation error observed at $\alpha \approx 0.1\%$ in \autoref{tab:Single-step}.

As shown in \autoref{tab:obs_noisy}, the assimilation errors for all variables increase with noise level $\tilde{\sigma}_{\text{noise}}$ as expected.
Importantly, FlowDA remains robust to observational perturbations, consistently reducing errors relative to the background across all variables. 
Notably, under extremely sparse observations ($\alpha \approx 0.1\%$), FlowDA on significantly noisy observations ($\tilde{\sigma}_{\text{noise}}=0.20$) still outperforms the noise-free DiffDA for most variables.

Notably, \autoref{tab:obs_noisy} indicates that distinct combinations of $(\alpha,\Tilde{\sigma}_{\text{noise}})$ can result in comparable error magnitudes. As illustrated in \autoref{fig:casestudy}, both $(\alpha\approx0.1\%,
\Tilde{\sigma}_{\text{noise}}=0.20)$ and $(\alpha\approx3.9\%,\Tilde{\sigma}_{\text{noise}}=0.00)$ yield similar T2M RMSE, while exhibiting qualitatively different assimilation dynamics. The former underestimates the background uncertainty and consequently applies more conservative updates due to limited trustworthy observations, whereas the latter leverages a higher observation density under elevated noise, which can induce overconfident estimates.

\paragraph{Auto-regressive DA} 
We assess the analysis skill and stability of FlowDA in long-horizon auto-regressive cycling DA. 
Specifically, we conduct a 15-day cycling DA experiment, following the protocol of DiffDA and VAE-Var for a fair comparison. We perform a 2-day free-running MLWP starting from the ERA5  reanalysis, yielding the day-0 background field for the following prediction-analysis cycle every 6 hours. 

We evaluate FlowDA against several representative baselines: DiffDA, VAE-Var (the SOTA for this task), 3D-Var, and bilinear interpolation. 
Baseline results are collected from \citet{xiao2024fengwu4dvar}. In their configuration, DiffDA uses GraphCast as the MLWP and adapts it for MLDA, whereas both 3D-Var and VAE-Var rely on Fengwu.

\begin{figure*}[h!] 
    \centering
    \includegraphics[width=0.9\linewidth]{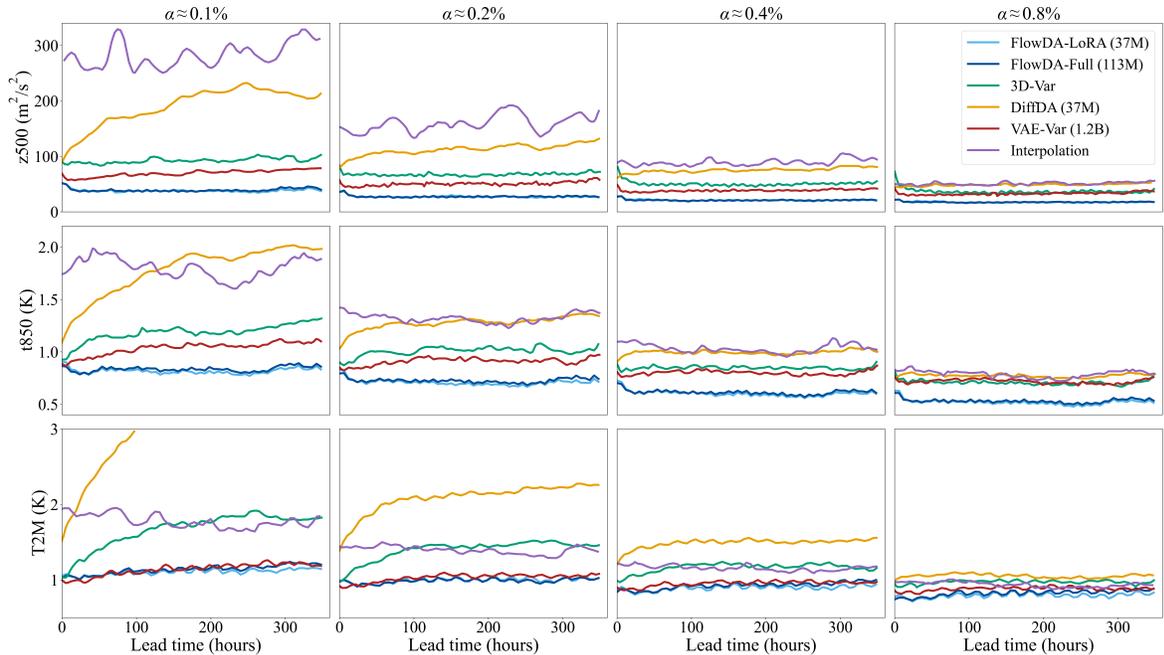}
    \caption{
    Benchmark comparison of FlowDA against baselines in a 15-day cycling DA  experiment (6-hour cycle) with four observation rates $\alpha$. For each rate, the observation locations are held fixed over the full cycle. 
    Shown are the lead-time evolutions of RMSE for z500, t850, and T2M relative to the ERA5 ground truth.
    }
    \label{fig:exp3_plot_fixed}
\end{figure*}

\begin{figure*}[h!]
    \centering
    \includegraphics[width=0.9\linewidth]{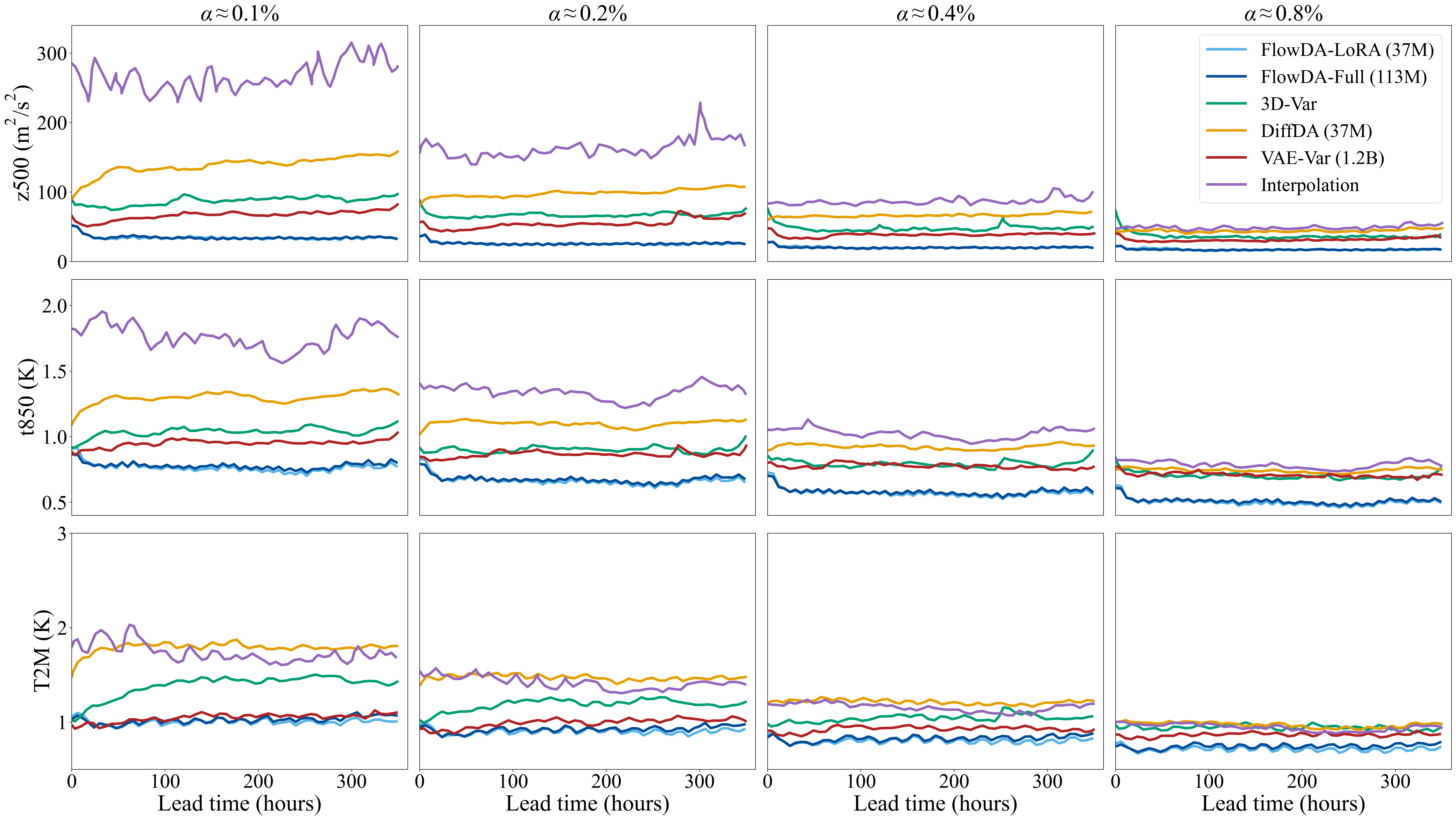}
    \caption{As in \autoref{fig:exp3_plot_fixed}, but with observation locations shuffled at every DA cycle.
    }
    \label{fig:exp3_plot_unfixed}
\end{figure*}

We consider two observation-location settings: (i) fixed locations throughout the 15-day period and (ii) locations shuffled (resampled) at every DA cycle. The resulting RMSE trajectories for z500, t850, and T2M are depicted in \autoref{fig:exp3_plot_fixed} and \autoref{fig:exp3_plot_unfixed}, respectively. As previously reported by DiffDA and VAE-Var, shuffled observations lead to lower errors across methods. Importantly, FlowDA remains the strongest performer in both regimes, as we discuss next.

As shown in \autoref{fig:exp3_plot_fixed}, FlowDA-LoRA attains the lowest error in most cases and closely matches FlowDA-Full, despite using only 37M trainable parameters---about one third of FlowDA-Full's trainable-parameter count and comparable to DiffDA.
While the analysis accuracy of DiffDA degenerates markedly as the observation rate $\alpha$ decreases, VAE-Var (the strongest baseline in our comparison) and FlowDA exhibit weak performance degeneration. More importantly, FlowDA evidently outperforms VAE-Var for z500 and t850 and and yields a modest gain for T2M. For z500 and t850 in particular, FlowDA at $\alpha\approx 0.1\%$ attains analysis errors comparable to VAE-Var using substantially more observations $\alpha\approx 0.39\%$.

\section{Conclusion and Discussion}

In conclusion, we present FlowDA, an accurate, low-latency weather data assimilation via conditional flow matching. We first convert sparse observations to a gridded model-space representation using SetConv layer, and then fine-tune the Aurora foundation model to learn a linear probability path from the background to the analysis conditioned on the observational estimated field.

We choose the ERA5 reanalysis as the training and test dataset, and construct synthetic observations by sampling a fraction (observation rate) of each snapshot. We benchmark FlowDA against baseline models on three classical DA settings: single-step DA with clean observations, that with synthetically perturbed observations, and auto-regressive forecast with cycling DA. Crucially, FlowDA achieves considerably lower analysis errors than existing baselines in most cases and, where comparable, provides substantially faster inference.

Effectively, FlowDA matches and, in some respects, surpasses the functionality of classical 3D-Var in meteorological DA via a data-driven generative ML paradigm. Incorporating temporally distributed observations, in a way analogous to 4D-Var, to achieve fully data-driven, time-aware assimilation remains a worth-pursuing direction. Moreover, real-world observations are often irregularly distributed and do not directly correspond to atmospheric state variables, underscoring the need for MLDA models capable of  both assimilation and observational preprocessing.

\section*{Acknowledgments}

Computation of the work was performed on resources of the National Supercomputing Centre, Singapore (https://www.nscc.sg), alongside those provided by NUS IT via a grant (NUSREC-HPC-00001).


\appendix
\section{Appendices}

\subsection{Notation}
\begin{table}[h!]
\centering 
\caption{Notation used in this study}
\begin{tabular}{lll}
\hline
Symbol & Meaning \\
\hline
$H$ & Number of latitude index in atmosphere state\\
$W$ & Number of longitude index in atmosphere state\\
$V$ & Number of meteorological variable index in atmosphere state\\
$M$ & Number of observation in each time\\
$\mathbf{y}_{t}$        & Observation at time $t$ \\
${\mathbf{x}}^{\mathrm{o}}_{t}$ & State estimate constructed from $\mathbf{y}_{t}$ at time $t$ \\

$\mathbf{x}_{t}$        & Atmosphere state at time $t$ \\
$\mathbf{x}^{\mathrm{b}}_{t}$      & Background (forecast) state at time $t$ \\
$\mathbf{x}^{\mathrm{a}}_{t}$      & Analysis state at time $t$ \\

$\mathcal{H}$           & Observation operator \\

$t$                     & Physical (real-world) time \\
$\tau$                  & Flow-matching pseudo-time \\

$\mathbf{u}^{\theta}_{\tau}$   & Flow-matching velocity field parameterized by $\theta$ \\

$h_{m}$   & Latitude of $m$-th observation \\
$h_{n}$   & Latitude of $n$-th state grid point \\
$w_{m}$   & Longitude of $m$-th observation \\
$w_{n}$   & Longitude of $n$-th state grid point \\
$\phi_{mn}$   & Weight measures the influence of the $m$-th observation on the $n$-th grid \\

$\bm{\rho}^{\mathrm{o}}_{t}$         & observation density distribution at time $t$ \\
$\alpha_m$                & local Observation rate around $m$-th observation\\
$\Tilde{\sigma}_{\text{noise}}$  & Dimensionless parameter indicating the magnitude of observational noises
\\

$\text{T2M}$                   & Temperature at 2 m \\
$\text{U10}$                   & Zonal wind at 10 m \\
$\text{V10}$                   & Meridional wind at 10 m \\
$\text{MSLP}$                  & Meridional wind at 10 m \\
$\text{q100}$                  & Specific humidity at 100 hPa \\
$\text{u300}$                  & Zonal wind at 300 hPa \\
$\text{v300}$                  & Meridional wind at 300 hPa \\
$\text{z500}$                  & Geopotential height at 500 hPa \\
$\text{t850}$                  & Temperature at 850 hPa \\

\hline
\end{tabular}
\end{table}

\section{Local observation rate}\label{sec:loc_obv} 
To reduce the computational cost, we adopt a $k$-neighbor SetConv variant~\citep{qi2017pointnetplus} that restricts weight computation to grid points among the $k$ nearest neighbors of each observation in both latitude and longitude. Under this setting, the local observation density $\alpha_m=M_k/(k\times k)$, where $M_k$ denotes the number of observations falling within this neighborhood. This density $\alpha_m$ is provided to the MLP kernel and combined with relative distances when computing the weights (see \autoref{sec:obs}). We choose $k=188$ in this work.

\end{document}